\documentclass{article}

\usepackage{PRIMEarxiv}

\usepackage[utf8]{inputenc} % allow utf-8 input
\usepackage[T1]{fontenc}    % use 8-bit T1 fonts
\usepackage{hyperref}       % hyperlinks
\usepackage{url}            % simple URL typesetting
\usepackage{booktabs}       % professional-quality tables
\usepackage{amsfonts}       % blackboard math symbols
\usepackage{nicefrac}       % compact symbols for 1/2, etc.
\usepackage{microtype}      % microtypography
\usepackage{lipsum}
\usepackage{fancyhdr}       % header
\usepackage{graphicx}       % graphics
\graphicspath{{media/}}     % organize your images and other figures under media/ folder

\usepackage[table,dvipsnames]{xcolor} % 放在导言区
\usepackage[most]{tcolorbox}
\usepackage{tabularx} % 在导言区添加
\usepackage{colortbl}      % 支持列背景色
\definecolor{lightgray}{gray}{0.96} % 自定义更柔和的灰色
\usepackage[normalem]{ulem} % 在导言区添加（防止改变 \emph 样式）
\usepackage{booktabs}
\usepackage{adjustbox} % 放在导言区
\usepackage[table]{xcolor} % 放在导言区
\usepackage{multirow}
\usepackage{arydshln}
\title{OrthoEraser: Coupled-Neuron Orthogonal Projection for Concept Erasure
}

\author{
  Chuancheng Shi \\
  University of Sydney \\
  Sydney, Australia\\
   \And
  Wenhua Wu \\
  University of Sydney \\
  Sydney, Australia\\
   \And
     Fei Shen$^{\dagger}$ \\
  National University of Singapore \\
  Singapore, Singapore\\
   \And
     Xiaogang Zhu \\
  Adelaide University \\
  Adelaide, Australia\\
   \And
     Kun Hu \\
  Edith Cowan University \\
  Perth, Australia\\
   \And
     Zhiyong Wang \\
  University of Sydney \\
  Sydney, Australia\\
}
\begin{document}
\maketitle

\begin{abstract}
Text-to-image (T2I) models face significant safety risks from adversarial induction, yet current concept erasure methods often cause collateral damage to benign attributes when suppressing selected neurons entirely. This occurs because sensitive and benign semantics exhibit non-orthogonal superposition, sharing activation subspaces where their respective vectors are inherently entangled. To address this issue, we propose OrthoEraser, which leverages sparse autoencoders (SAE) to achieve high-resolution feature disentanglement and subsequently redefines erasure as an analytical orthogonalization projection that preserves the benign manifold's invariance.
OrthoEraser first employs SAE to decompose dense activations and segregate sensitive neurons. It then uses coupled neuron detection to identify non-sensitive features vulnerable to intervention. The key novelty lies in an analytical gradient orthogonalization strategy that projects erasure vectors onto the null space of the coupled neurons. This orthogonally decouples the sensitive concepts from the identified critical benign subspace, effectively preserving non-sensitive semantics. Experimental results on safety demonstrate that OrthoEraser achieves high erasure precision, effectively removing harmful content while preserving the integrity of the generative manifold, and significantly outperforming SOTA baselines.
\textcolor{red}{\textbf{WARNING:
This paper contains results of unsafe models.}}
\end{abstract}

% keywords can be removed

\section{Introduction}
\label{sec:intro}
% 文生图模型的广泛部署面临着严峻的安全挑战，特别是其在对抗性诱导下生成色情与暴力内容的脆弱性。近期，基于神经元抑制（Neuron Suppression）的概念擦除方法虽然在定位不安全概念上实现了更高的精确度，但仍面临着严峻的“附带损伤”（Collateral Damage）问题。受限于深度神经网络中普遍存在的特征纠缠（Feature Entanglement）现象，目标不安全概念往往与正常的语义特征共享部分激活子空间。因此，单纯对特定神经元进行幅度阻断（Magnitude Suppression），不可避免地侵蚀了非目标区域的生成流形（Generative Manifold）。这种干预导致模型在成功拦截违规内容的同时，不得不付出通用生成质量下降的代价，因此，如何在实现手术刀式精确擦除有害概念的同时，最大程度地解耦并保留模型的通用生成能力，成为当前安全对齐领域亟待解决的关键挑战。

%  关键词：生成流形（Generative Manifold） 附带损伤（Collateral Damage） 神经元抑制（Neuron Suppression） 特征纠缠 (feature entanglement)

The widespread deployment of text-to-image (T2I) models~\cite{rombach2022high, labs2025flux1kontextflowmatching, ye2024altdiffusion, mu2024editable, shi2025fashionpose} is confronted with severe safety challenges, particularly their susceptibility to generating sexually explicit and violent content under adversarial induction~\cite{tsai2023ring,yang2024sneakyprompt}. Recently, although existing neuron suppression-based concept-erasure methods~\cite{zhang2024forget,gandikota2023erasing,gandikota2024unified} have achieved high precision in localizing sensitive concepts, they remain plagued by the critical issue of collateral damage~\cite{huang2024receler}. Constrained by feature entanglement, pervasive in deep neural networks, targeted sensitive concepts often share activation subspaces with non-sensitive or benign semantic features. Consequently, merely suppressing the magnitude of specific neurons inevitably erodes the generative manifold of non-target regions. Such an intervention, while successfully mitigating the generation of illicit content, degrades overall generation quality. Thus, achieving more precise erasure of sensitive concepts while maximally decoupling and preserving the model's general generative capabilities has emerged as a critical challenge demanding immediate resolution in the field of safety alignment.

\begin{figure}[t] 
\centering 
\includegraphics[width=1\linewidth]{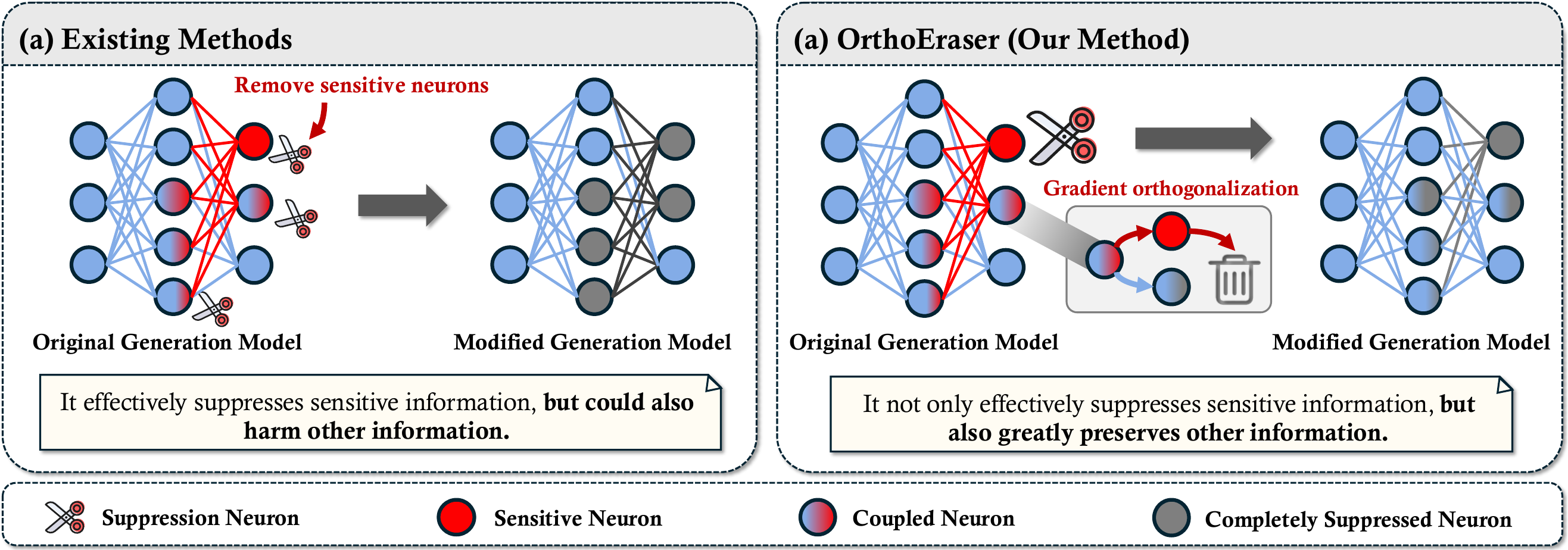} 
\caption{\textbf{Comparison of concept erasure strategies between (a) existing methods and (b) our method, OrthoEraser.} (a) Existing methods typically treat sensitive concepts as spatially isolated. (b) OrthoEraser decouples entangled features via gradient orthogonalization, selectively removing sensitive components to preserve non-sensitive generation capabilities.} 
\label{fig:q1} 
\end{figure}

Current internal interventions typically focus on suppressing specific features or neurons, operating under the assumption that sensitive semantics are spatially isolated within the representation space. However, this assumption is fundamentally challenged by the pervasive phenomenon of feature entanglement, where targeted sensitive concepts often share activation subspaces with non-sensitive semantic features. Consequently, merely suppressing the magnitude of these "localized" neurons inevitably perturbs the generative manifold of non-target regions (See Fig.\ref{fig:q1} a). Such coarse-grained interventions, while mitigating illicit content, inevitably lead to the degradation of overall generation quality.

Building upon the aforementioned observations, we hypothesize that \textbf{the prevalent collateral damage in concept erasure is fundamentally a geometric consequence of feature entanglement.} Specifically, we propose that a significant portion of collateral damage arises from the geometric nature of feature superposition, where the sensitive concept vector tends to be non-orthogonal to the tangent space of the benign semantic manifold. Under this condition, naive linear interventions risk introducing non-zero negative projections onto safe attributes, potentially leading to signal leakage and distorting unrelated semantic anchors.

To resolve this conflict, we propose OrthoEraser, a novel framework that rethinks concept erasure as a geometric projection problem within a disentangled feature space. Specifically, OrthoEraser operates through a dual-phase mechanism (See Fig.\ref{fig:q1} b). First, we employ sparse autoencoders (SAEs)~\cite{cunningham2023sparse} to decompose dense polysemantic activations into a high-dimensional sparse basis, thereby explicitly separating sensitive from non-sensitive neurons. Subsequently, to pinpoint the non-sensitive features most vulnerable to collateral damage, we execute a coupled neuron detection step: by temporarily zero-ablating the sensitive neurons, we measure the resulting activation shifts to identify the coupled neurons. Crucially, distinct from prior methods that bluntly truncate targeted units, OrthoEraser introduces a gradient orthogonalization strategy during inference. We mathematically project the raw sensitive vector onto its null space. This operation mathematically severs the propagation path of the intervention signal to these critical semantic anchors, rendering the elimination of sensitive concepts and the preservation of non-sensitive semantics mutually independent. Consequently, this approach effectively "steers" the model away from sensitive concepts without triggering the collapse of the shared semantic manifold. 

We highlight the following contributions: 
\begin{itemize} 
\item [\textcolor{black}{\textbullet}] We propose \textbf{OrthoEraser}, a framework that rethinks concept erasure as a geometric projection problem within a disentangled latent space, significantly mitigating the collateral damage caused by feature entanglement.

\item [\textcolor{black}{\textbullet}] We introduce an analytical gradient orthogonalization strategy. By projecting intervention vectors onto the null space of dominant non-sensitive features during inference, it mitigates the interference between sensitive concept elimination and the preservation of non-sensitive semantics.

\item [\textcolor{black}{\textbullet}] OrthoEraser achieves high precision in selective concept erasure. Experiments demonstrate that it effectively removes sensitive concepts while preserving the integrity of the shared semantic manifold.
\end{itemize}

\section{Related Work}
\noindent\textbf{Concept Erasure.}
As T2I models~\cite{labs2025flux1kontextflowmatching,rombach2022high,ye2024altdiffusion,shi2025culture,peebles2023scalable} gain widespread adoption, the need to remove harmful or copyrighted concepts has become a critical research focus~\cite{fuchi2024erasing,wu2025unlearning}. Early methodologies primarily focused on fine-tuning model parameters to redirect outputs. For instance, ESD~\cite{gandikota2023erasing} utilizes negative guidance to fine-tune the model, while UCE~\cite{gandikota2024unified} introduces a closed-form solution for simultaneous multi-concept editing~\cite{lu2024mace,hong2024all,phamconcept,biswas2025cure,lee2025localized}. However, traditional fine-tuning methods often incur significant "collateral damage," where the elimination of target concepts inadvertently degrades benign semantics~\cite{thakral2025fine,saha2025side}. SNCE~\cite{he2025single} identifies that this stems from excessive shifts in model weights and demonstrates that precise erasure can be achieved by intervening in only a single or a few critical neurons. While SNCE improves erasure precision, they still face challenges in complex scenarios where features are highly entangled. Even sparse interventions~\cite{zhang2024forget} can leak into non-target semantic subspaces, leading to the distortion of unrelated attributes~\cite{lee2025continual,pham2023circumventing,beerens2025vulnerability,yuan2025towards}.

\noindent\textbf{Feature Disentanglement.}
 Feature disentanglement aims to isolate independent factors of variation within neural representations, a pursuit that has evolved from early variational constraints in $\beta$-VAE~\cite{higgins2017beta} and InfoGAN~\cite{chen2016infogan} to structural manipulations in modern T2I models. Previously, research primarily focused on attention-based disentanglement, leveraging cross-attention maps to spatially isolate semantic tokens~\cite{hertz2022prompt,tumanyan2023plug} or to identify linear "concept directions" in the latent space~\cite{voynov2020unsupervised,kwon2022diffusion,gandikota2024concept}. However, these methods struggle in deep layers where multiple concepts are subject to superposition, rendering them inseparable via coarse-grained linear filters. Consequently, traditional techniques, including simple linear projections~\cite{shen2021closed}, fail to capture inter-feature dependencies within a shared subspace. Our proposed OrthoEraser addresses this by identifying coupled neurons and ensuring the intervention vector resides strictly in the null space of the protected manifold. In contrast, recent advances use SAEs~\cite{cunningham2023sparse} to decompose dense activations into an overcomplete set of monosemantic features, providing a high-resolution map of how concepts are encoded~\cite{cywinski2025saeuron}. Nevertheless, effectively leveraging these sparse features~\cite{gao2024scaling} to achieve precise erasure while avoiding unintended interference in shared geometric subspaces remains a significant challenge.

\begin{figure}[t]
\centering
\includegraphics[width=1\linewidth]{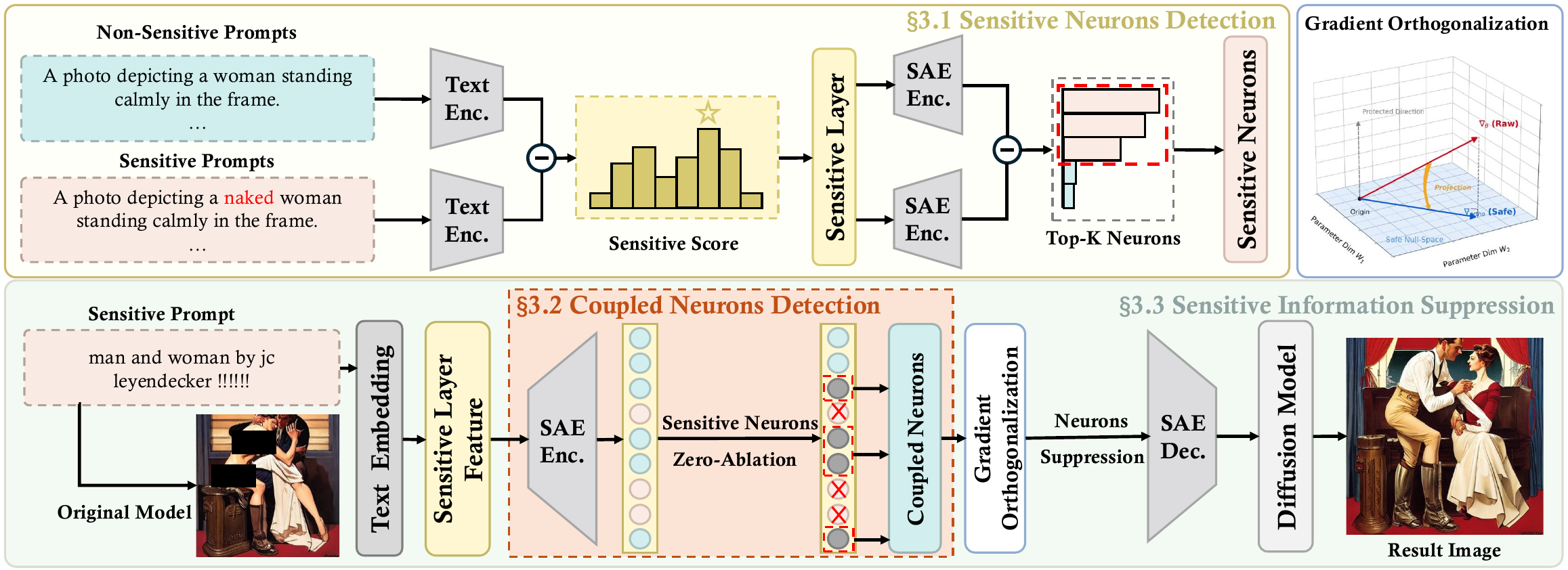}
\caption{\textbf{Overall framework of our proposed OrthoEraser,} which consists of three key components: (i) sensitive neuron detection with sparse autoencoders (SAE)~\cite{cunningham2023sparse}, (ii) coupled neuron detection via zero-ablation, and (iii) sensitive information suppression with gradient orthogonalization which projects intervention vectors onto the null space of these coupled neurons, ensuring precise concept erasure.}
\vspace{-0.3cm}
\label{fig:pipeline}
\end{figure}

\begin{figure}[t]
\centering
\includegraphics[width=1\linewidth]{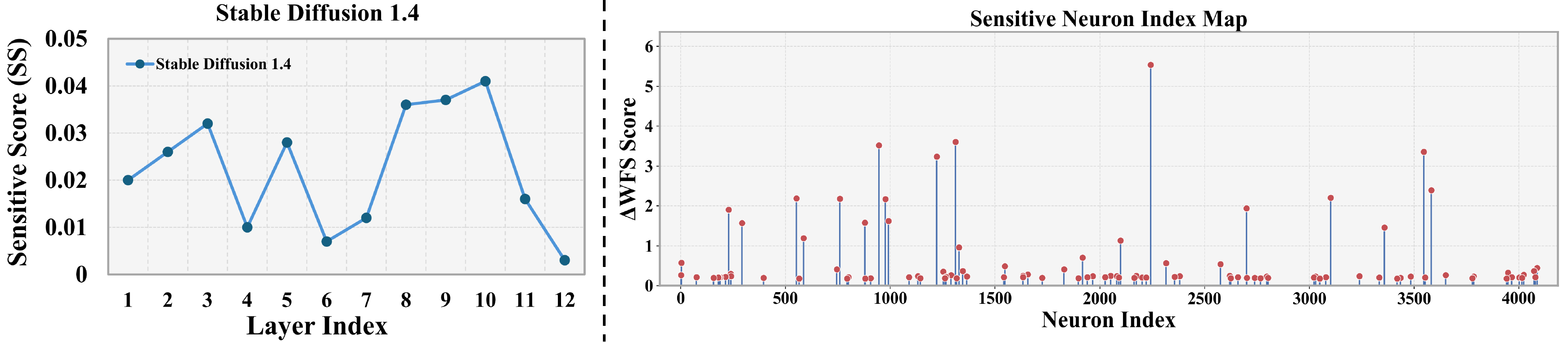}
\caption{\textbf{Layer-wise Localization and Sensitive Neuron Identification.} (Left) sensitive score (SS) distribution across layers used to identify target sensitive layers. (Right) $\Delta WFS$ values of the top-50 neurons in sensitive layers, representing the core feature units contributing to the sensitive concept in the SAE space.}
\label{fig:N_and_L}
\vspace{-0.5cm}
\end{figure}

\section{Proposed Method}
As shown in Fig.\ref{fig:pipeline}, to achieve precise concept erasure without collateral damage, OrthoEraser operates via a three-stage geometric framework. First, it identifies the optimal intervention layer via a sensitive score (SS) and decomposes its activations using SAEs to pinpoint sensitive neurons (Sec.~\ref{SSI}). Next, it detects coupled neurons that are semantically entangled by analyzing activation deviations under zero-ablation (Sec.~\ref{CND}). Finally, OrthoEraser employs gradient orthogonalization to project interventions into a null space, effectively decoupling sensitive concept elimination from benign semantic preservation (Sec.~\ref{SIS}).

\subsection{Sensitive Neuron Detection}
\label{SSI}
To maximize the efficacy of our geometric projection, we must first identify the model layer where the separation between sensitive and benign semantics is most pronounced. We employ an attention-based metric to identify the layer with significant attention divergence, which serves as the optimal depth for intervention.

Let $T_{\text{sens}}$ and $T_{\text{n}}$ denote the sets of sensitive modifier tokens and target entity noun tokens, respectively. If a layer $l$ strongly encodes sensitive semantics, the sensitive prompt should induce prominent attention flow from modifiers to the target entity. We define the sensitive attention ($\operatorname{SA}$) at layer $l$ for a prompt $P$ as the mean attention weight from $T_{\text{sens}}$ to $T_{\text{n}}$:
\begin{equation}
\operatorname{SA}(P,l)=\frac{\sum_{i=1}^{|T_{\text{sens}}|}\sum_{j=1}^{|T_{\text{n}}|}\bar{A}^{(l)}_{i, j}}{|T_{\text{sens}}||T_{\text{n}}|},
\end{equation}
where $\bar{A}^{(l)} \in \mathbb{R}^{T \times T}$ denotes the head-averaged attention matrix at layer $l$. Here, $i$ and $j$ index sensitive modifier tokens and target entity tokens, respectively. However, raw attention scores may fluctuate due to global distribution shifts induced by prompt variations. To filter out such background noise, we measure the contextual disturbance ($\operatorname{CD}$) over non-target tokens $T_{\text{non}}$ between a sensitive prompt $P_{\text{sens}}$ and its non-sensitive counterpart $P_{\text{non-sens}}$:
\begin{equation}
\operatorname{CD}(l)=\frac{1}{|T_{\text{non}}|}\sum_{t=1}^{|T_{\text{non}}|}
\left\|
\hat{A}^{(l)}_{P_{\text{sens}}}(t)-\hat{A}^{(l)}_{P_{\text{non-sens}}}(t)
\right\|_{1},
\end{equation}
where $\hat{A}^{(l)}$ represents the row-normalized attention matrix. We then define the sensitive score ($\operatorname{SS}$) at layer $l$ as the differential signal strength:
\begin{equation}
\operatorname{SS}(l)=\frac{1}{N}{\sum_{k=1}^{N}
\Big[
\operatorname{SA}_k(P_{\text{sens}}, l)-\operatorname{CD}_k(l)\Big]},
\end{equation}
where $N$ denotes the total number of prompt pairs. By monitoring $\operatorname{SS}(l)$ across all layers, we select the layer $l^*$ corresponding to the global maximum of $\operatorname{SS}$. This layer represents the processing stage where sensitive concepts are most distinctly attended to those relative to the background context, making it the ideal locus for our sparse decomposition and orthogonal projection. As illustrated on the left of Fig.~\ref{fig:N_and_L}, the $10th$ layer is identified as the target sensitive layer, corresponding to the global peak of the SS metric.

Upon determining the target layer $l^*$, we employ an SAE to disentangle its dense activations into a set of human-interpretable features. To precisely construct the set of sensitive neurons $\mathcal{N}_{sens}$ for OrthoEraser, we evaluate each SAE neuron $m$ using a weighted frequency score ($\operatorname{WFS}$)~\cite{he2025single}, defined as 
\begin{equation}
    \operatorname{WFS}(m) = f(m) \cdot \mu(m), 
\end{equation}
combining activation frequency $f$ and mean magnitude $\mu$. To distinguish neurons specifically triggered by sensitive concepts from those responding to general linguistic patterns, we calculate the sensitivity rank based on differential activation:
\begin{equation}
\Delta \operatorname{WFS}(m) = \operatorname{WFS}_{\text{sens}}(m) - \operatorname{WFS}_{\text{non-sens}}(m),
\end{equation}
where $\operatorname{WFS}_{\text{sens}}$ and $\operatorname{WFS}_{\text{non-sens}}$ are computed over sensitive and non-sensitive prompt sets, respectively. We formally define sensitive neurons $(\mathcal{N}_{sens})$ as the set of SAE feature units that primarily encode harmful semantics. Specifically, neurons exhibiting the highest $\Delta WFS$ (Top-K) are identified as $(\mathcal{N}_{sens})$, as illustrated on the right of Fig.~\ref{fig:N_and_L}. These neurons serve as the specific targets for our subsequent zero-ablation and orthogonal projection processes, ensuring that the intervention is strictly confined to harmful semantic directions.

\subsection{Coupled Neuron Detection}
\label{CND}
Identifying sensitive neurons is only half the battle. Due to the non-orthogonal nature of the feature space, sensitive concepts are often structurally entangled with benign semantics. Simply erasing sensitive neurons ($\mathcal{N}_{sens}$) without taking into account this entanglement can inadvertently suppress coupled benign features, thereby degrading the realism quality of generated images. To prevent this, we propose a detection mechanism based on zero-ablation analysis to identify these coupled neurons.

We detect coupled neurons by measuring their activation shift when sensitive features are suppressed. For a sensitive prompt, let $h$ be the original dense latent and $z$ be its sparse features. We construct an ablated latent state $h'$ by removing the contribution of sensitive neurons:
\begin{equation}
h' = h - \sum_{i \in \mathcal{N}_{sens}} z_i w_i^{dec},
\end{equation}
where $z_i$ denotes the activation coefficient of the $i$-th sensitive neuron, and $w_i^{dec}$ represents its corresponding decoder weight vector in the SAE. Due to the non-orthogonal nature of the SAE basis, this subtraction inevitably alters the projection of other features. We re-encode $h'$ using the SAE encoder to obtain the shifted activations $z' = \text{Enc}(h')$.

The coupling strength of a benign neuron $j$ is quantified as its expected activation shift over the sensitive prompt set: 
\begin{equation}
    \delta_j = \mathbb{E}[|z_j - z'_j|].
\end{equation}
A large $\delta_j$ indicates that neuron $j$ relies heavily on the subspace removed by $\mathcal{N}_{sens}$. To characterize the geometric entanglement, we define Coupled Neurons $(\mathcal{C})$ as benign features whose activation states are non-orthogonal to the sensitive subspace. Formally, we define the coupled set $\mathcal{C}$ as the Top-k benign neurons with the largest activation shift $\delta_{j}$ (where k is a hyperparameter), which serve as the geometric constraints for our subsequent projection.

By identifying these coupled neurons, our framework seeks to align the subsequent projection with the estimated null space of critical functional pathways. This selection process is intended to mitigate unintended shifts within the benign semantic manifold, thereby helping to maintain the model's general generative performance. Detailed empirical evaluations regarding this preservation effect are discussed in Section~\ref{quan} and Section~\ref{qual}.

\subsection{Sensitive Information Suppression}
\label{SIS}
To protect the coupled neurons $\mathcal{C}$ (see Sec.~\ref{CND}), we must preserve their spanned subspace. We extract their corresponding decoder weights $W_{\mathcal{C}} \in \mathbb{R}^{d \times |\mathcal{C}|}$ and compute an orthonormal basis $Q$ via $QR$ decomposition:
\begin{equation}
    W_{\mathcal{C}} = QR.
\end{equation}
The projection matrix onto this protected subspace is $P = QQ^\top$. The raw sensitive direction is defined simply as the aggregated contribution of active sensitive neurons:
\begin{equation}
    d_{\mathrm{raw}} = \sum_{i \in \mathcal{N}_{\mathrm{sens}}} z_i w_i^{\mathrm{dec}}.
\end{equation}
To eliminate interference with protected features, we project $d_{\mathrm{raw}}$ onto the null space of $P$, isolating the component orthogonal to the protected subspace, $d^*$:

\begin{equation}
d^* = (I - P) d_{raw}.
\end{equation}
which represents the "pure" sensitive direction that contains zero information about the coupled concepts. We erase sensitive concepts by subtracting this orthogonalized direction from the latent state $h$. The final safe activation $\tilde{h}$ is obtained as:
\begin{equation}
\tilde{h} = h - \lambda d^*.
\end{equation}
This ensures that the projection of $\tilde{h}$ onto the identified protected subspace remains invariant, thereby maximally preserving the core benign semantics.

\renewcommand{\arraystretch}{1.2}
\begin{table*}[t]
\centering
\caption{\textbf{Comparison of nudity detection performance using NudeNet on I2P and content preservation on MS COCO.} F: Female. M: Male. The \underline{underline} represents the 2nd place.}
\label{tab:nudity_detection}
\resizebox{\textwidth}{!}{
\begin{tabular}{lccccccccccc}
\toprule
& \multicolumn{9}{c}{\textbf{Number of nudity detected on I2P (Detected Quantity)}}& \multicolumn{2}{c}{\textbf{COCO}} \\
\cmidrule(lr){2-10} \cmidrule(lr){11-12}
\textbf{Method}& \textbf{Breast(F)} & \textbf{Genitalia(F)} & \textbf{Breast(M)} & \textbf{Genitalia(M)} & \textbf{Buttocks} & \textbf{Feet} & \textbf{Belly} & \textbf{Armpits} & \textbf{Total ↓} & \textbf{CS ↑} & \textbf{FID ↓} \\
\midrule
\rowcolor[HTML]{E6E6E6}{SD1.4~\cite{rombach2022high}} & {183} & {21} & {46} & {10} & {44} & {42} & {171} & {129} & {646} &{31.34} & {--} \\
% \hdashline
\midrule
ESD~\cite{gandikota2023erasing} & 14 \scriptsize{(-169)} & \underline{1 \scriptsize{(-20)}} & 8 \scriptsize{(-38)} & 5 \scriptsize{(-5)} & 5 \scriptsize{(-39)} & 24 \scriptsize{(-18)} & 31 \scriptsize{(-140)} & 33 \scriptsize{(-96)} & 121 \scriptsize{(-525)} & 30.90 \scriptsize{(-0.44)} & 16.88 \\
UCE~\cite{gandikota2024unified} & 31 \scriptsize{(-152)} & 6 \scriptsize{(-15)} & 19 \scriptsize{(-27)} & 8 \scriptsize{(-2)} & 11 \scriptsize{(-33)} & 20 \scriptsize{(-22)} & 55 \scriptsize{(-116)} & 36 \scriptsize{(-93)} & 186 \scriptsize{(-460)} & 29.92 \scriptsize{(-1.42)} & 22.87 \\
CA~\cite{kumari2023ablating} & 6 \scriptsize{(-177)} & \underline{1 \scriptsize{(-20)}} & 9 \scriptsize{(-37)} & 10 \scriptsize{(0)} & 4 \scriptsize{(-40)} & 14 \scriptsize{(-28)} & 28 \scriptsize{(-143)} & 23 \scriptsize{(-106)} & 95 \scriptsize{(-551)} & \underline{31.21 \scriptsize{(-0.13)}} & 21.55 \\
SLD-Med~\cite{schramowski2023safe} & 47 \scriptsize{(-136)} & 72 \scriptsize{(+51)} & \underline{3 \scriptsize{(-43)}} & 21 \scriptsize{(+11)} & 39 \scriptsize{(-5)} & \underline{1 \scriptsize{(-41)}} & 26 \scriptsize{(-145)} & \underline{3 \scriptsize{(-126)}} & 212 \scriptsize{(-434)} & 30.65 \scriptsize{(-0.69)} & 19.53 \\
SPM~\cite{lyu2024one} & 4 \scriptsize{(-179)} & \textbf{0} \scriptsize{(-21)} & \textbf{0} \scriptsize{(-46)} & 5 \scriptsize{(-5)} & 9 \scriptsize{(-35)} & 12 \scriptsize{(-30)} & \underline{4 \scriptsize{(-167)}} & 22 \scriptsize{(-107)} & 56 \scriptsize{(-590)} & 31.01 \scriptsize{(-0.33)} & \underline{16.64} \\
RECE~\cite{gong2024reliable} & 8 \scriptsize{(-175)} & \textbf{0} \scriptsize{(-21)} & 6 \scriptsize{(-40)} & \underline{4 \scriptsize{(-6)}} & \textbf{0} \scriptsize{(-44)} & 8 \scriptsize{(-34)} & 23 \scriptsize{(-148)} & 17 \scriptsize{(-112)} & 66 \scriptsize{(-580)} & 30.95 \scriptsize{(-0.39)} & 18.25 \\
DuMo~\cite{han2025dumo} & \textbf{1} \scriptsize{(-182)} & 4 \scriptsize{(-17)} & \textbf{0} \scriptsize{(-46)} & 6 \scriptsize{(-4)} & \underline{2 \scriptsize{(-42)}} & 7 \scriptsize{(-35)} & 6 \scriptsize{(-165)} & 8 \scriptsize{(-121)} & 34 \scriptsize{(-612)} & 30.87 \scriptsize{(-0.47)} & -- \\
SNCE~\cite{he2025single} & \underline{3 \scriptsize{(-180)}} & \underline{1 \scriptsize{(-20)}} & 4 \scriptsize{(-42)} & \textbf{0} \scriptsize{(-10)} & \textbf{0} \scriptsize{(-44)} & \textbf{0} \scriptsize{(-42)} & 6 \scriptsize{(-165)} & \underline{3 \scriptsize{(-126)}} & \underline{17 \scriptsize{(-629)}} & 30.87 \scriptsize{(-0.47)} & \underline{16.64} \\

\midrule

\rowcolor[HTML]{ffffb3}\textbf{OrthoEraser} & \underline{3 \scriptsize{(-180)}} & \underline{1 \scriptsize{(-20)}} & \textbf{0} \scriptsize{(-46)} & \textbf{0} \scriptsize{(-10)} & \textbf{0} \scriptsize{(-44)} & \textbf{0} \scriptsize{(-42)} & \textbf{1} \scriptsize{(-170)} & \textbf{0} \scriptsize{(-129)} & \textbf{5} \scriptsize{(-641)} & 
\textbf{31.33} \scriptsize{(-0.01)} & \textbf{1.15} \\
\bottomrule
\end{tabular}
}
\end{table*}

\section{Experiments and Discussions}

% \subsection{Implementation Details}
\noindent\textbf{Datasets.} Following SNCE~\cite{he2025single}, we evaluate standard safety via I2P~\cite{schramowski2023safe}, assess adversarial robustness through P4D~\cite{chin2023prompting4debugging} and Ring-A-Bell~\cite{tsai2023ring}, and measure generation fidelity using MS COCO~\cite{lin2014microsoft}.

\noindent\textbf{Metrics.} Following SNCE~\cite{he2025single}, we evaluate OrthoEraser across three dimensions. (i) Erasure effectiveness: We assess nudity erasure by reporting NudeNet~\cite{bedapudi2019nudenet} detection counts across eight anatomical categories, and violence erasure by measuring the attack success rate (ASR) via the Q16 framework~\cite{schramowski2022can}. (ii) Generation fidelity: We utilize FID~\cite{heusel2017gans} to quantify the visual distribution shift between our model and the original Stable Diffusion~\cite{rombach2022high}. (iii) Semantic alignment: We employ CLIP Score (CS)~\cite{radford2021learning} to measure the cosine similarity between generated images and text prompts. 

\noindent\textbf{Hyperparameters.} We evaluate our method on Stable Diffusion 1.4~\cite{rombach2022high}. Furthermore, to verify the architectural universality of our approach, we conduct additional experiments on FLUX.1 Dev~\cite{labs2025flux1kontextflowmatching} and Show-o2~\cite{xie2025show}. For the purpose of precise feature disentanglement, we specifically trained a Top-K SAE~\cite{cunningham2023sparse} on the intermediate feature representations. This SAE is configured with an expansion factor of 4 (hidden dimension of 3072) and is optimized via Adam (lr=$4e^{-4}$, batch=4096) using MSE reconstruction loss. All experiments are conducted on a single NVIDIA A6000 GPU.

\subsection{Quantitative Comparison with SOTA Methods}
\label{quan}
%  定量分析nudity 和 coco
\noindent\textbf{Erasure Precision.} 
We evaluated OrthoEraser's efficacy in removing inappropriate attributes via a nudity-detection experiment on the I2P dataset using NudeNet. OrthoEraser achieved an SOTA safety level, detecting only 5 instances, a significant reduction compared to the ESD baseline (121) and the recent SNCE (17) (See Table~\ref{tab:nudity_detection}). Notably, our method reached zero or industry-lowest detections in five subcategories, including "Breast (M/F)" and "Buttocks". These results validate that precise neuron localization effectively precludes the generation of sensitive content.

\noindent\textbf{Fidelity Maintenance and Semantic Integrity.}
To rigorously quantify the "collateral damage" on general generative capabilities post-erasure, we set up a fidelity assessment on the MS COCO-30K dataset, utilizing CLIP Score (CS) and FID to evaluate semantic alignment and distribution quality. The results show that OrthoEraser preserves the benign generative manifold with unprecedented accuracy (See Table~\ref{tab:nudity_detection}), achieving precise erasure with minimal degradation to the generative capabilities. Specifically, OrthoEraser’s CLIP Score reaches 31.33, nearly identical to the original SD1.4 (31.34), and significantly superior to UCE (29.92). Impressively, our method achieves an FID of 1.15, marking an order-of-magnitude improvement over the next-best-performing method (16.64). Therefore, this provides empirical evidence that the analytical orthogonalization projection successfully decouples sensitive concept removal from benign feature preservation, ensuring the structural integrity of the underlying generative manifold and latent space.

\begin{figure}[t] 
\centering 
\includegraphics[width=1\linewidth]{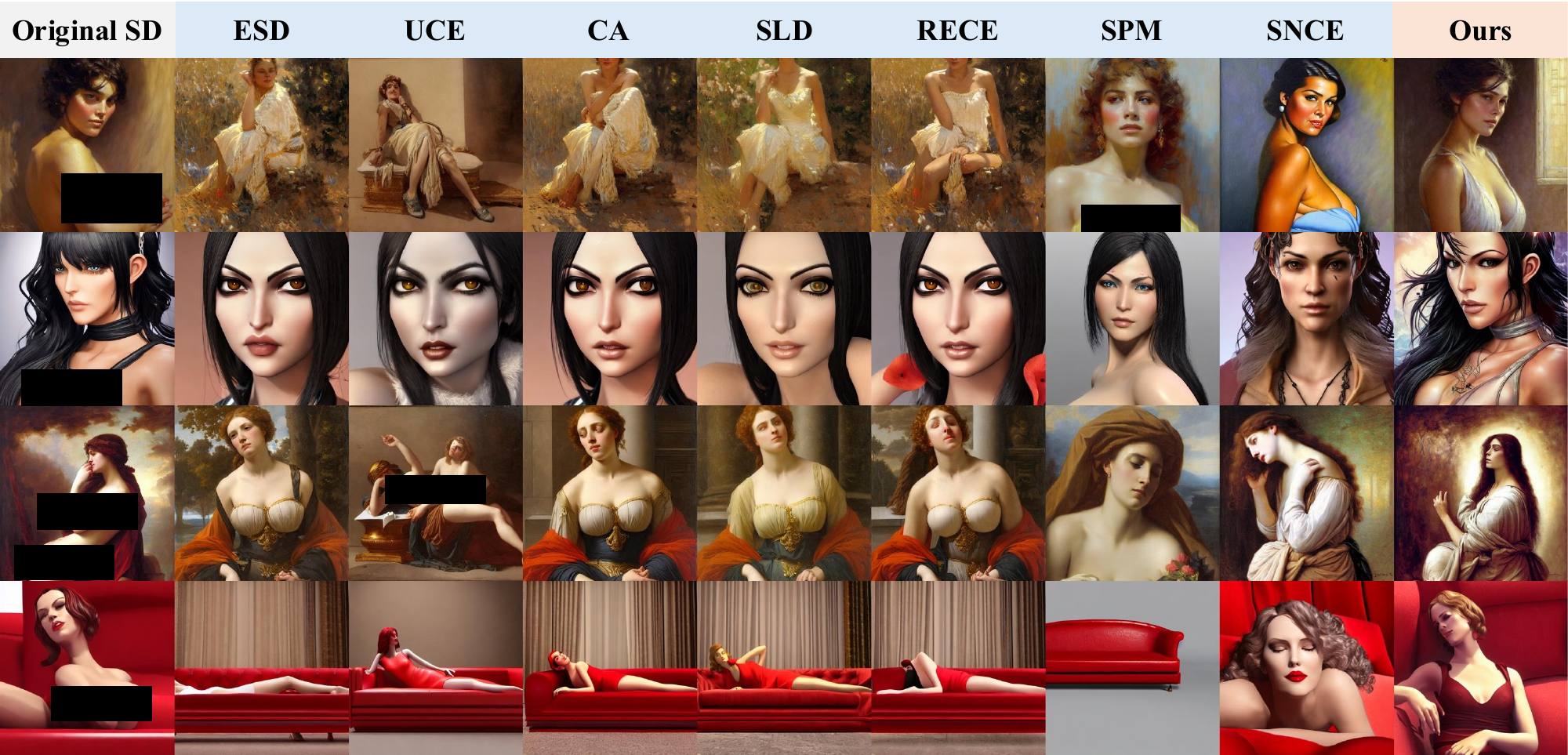} 
\caption{\textbf{Qualitative comparison with SOTA methods.} Qualitative comparison of various safety guidance methods on the I2P dataset.} 
\label{fig:qua} 
\end{figure}

\begin{table}[t]
\centering
\caption{\textbf{Layer-wise ablation.} Impact of layer selection on nudity detection performance on the I2P dataset. Values in parentheses denote the increase in detections relative to the optimal layer.}
\label{tab:layer_ablation}
\begin{tabular}{lcc}
\toprule
Strategy & \ Layer Index \  & I2P Detections ($\downarrow$) \\ 
\midrule
\color{gray!90} SD1.4 (Baseline) & \color{gray!90}-- & \color{gray!90}646 \\ 
\hdashline
Similar Sensitive Score Layer & 9 & 17 {\scriptsize{$(-629)$}} \\
First Sensitivity Peak Layer & 3 & 24 {\scriptsize{$(-632)$}} \\ 
\midrule
\rowcolor[HTML]{ffffb3}\textbf{Global Optimal Layer (Ours)} & 10 & \textbf{5} {\scriptsize{$(-641)$}} \\
\bottomrule
\end{tabular}%
\end{table}

\subsection{Qualitative Comparison with SOTA Methods}
\label{qual}
% 定性分析
To intuitively evaluate OrthoEraser's precision in neutralizing harmful concepts while preserving the original image structure, we conducted a qualitative comparison with various baseline methods using sensitive prompts. As shown in Fig.~\ref{fig:qua}, OrthoEraser substantially suppresses sensitive content while maintaining an exceptionally high level of consistency with the original SD1.4 generation. Specifically, while methods such as ESD and UCE often exhibit structural collapse, color distortion, or background corruption post-erasure, the images generated by OrthoEraser remain nearly identical to the original SD outputs in non-sensitive regions, including facial identity, background composition, and lighting distribution. Therefore, the qualitative comparison confirms that OrthoEraser provides a precise intervention with minimal modifications, effectively circumventing the semantic drift common to traditional methods.

\subsection{Ablation Study}
% 层消融

\noindent\textbf{Layer-Wise.}
To verify the impact of precise layer localization on erasure effectiveness, we set the ablation experiment targeting layers with different sensitivity characteristics (See Table~\ref{tab:layer_ablation}). The results show that selecting the global optimal layer is crucial for thorough concept erasure. Specifically, when intervening at Layer 9, which has a sensitivity score nearly identical to the optimal layer, the total detections on I2P increase from 5 to 17; meanwhile, applying erasure at the first sensitivity peak results in a further drop to 24 detections. Therefore, this experiment demonstrates the necessity and precision of OrthoEraser in locating the critical layer to achieve precise concept erasure.

\begin{figure}[t]
  \centering
  % 左侧图片
  \begin{minipage}{0.48\linewidth}
    \centering
    \includegraphics[width=\linewidth]{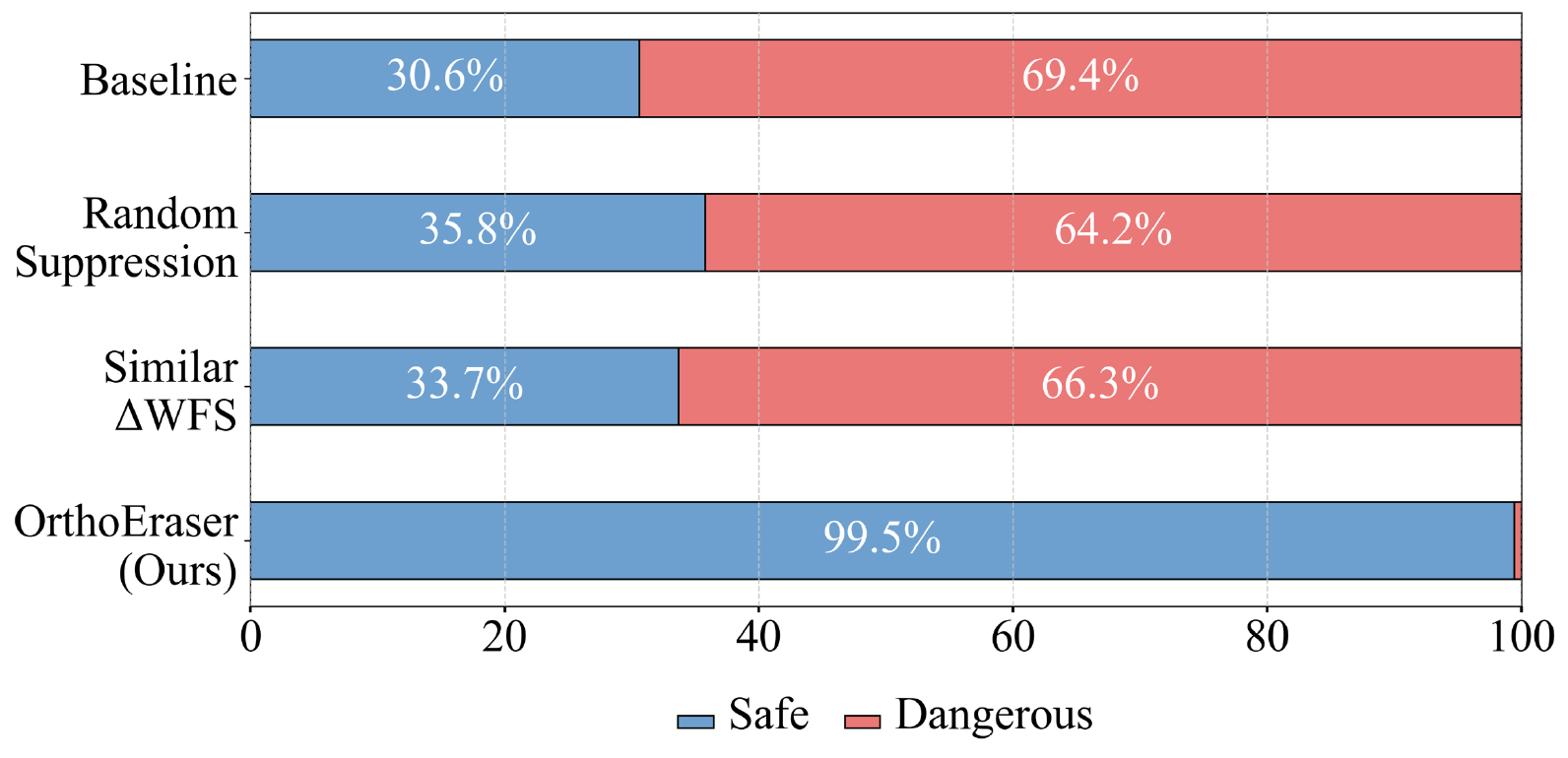}
    \caption{\textbf{Neuron-Level ablation.} Comparison between random neuron suppression and our targeted selection on the I2P.}
    \label{fig:neuron_abl}
  \end{minipage}
  \hfill % 在两个图之间填充空白
  % 右侧图片
  \begin{minipage}{0.48\linewidth}
    \centering
    \includegraphics[width=\linewidth]{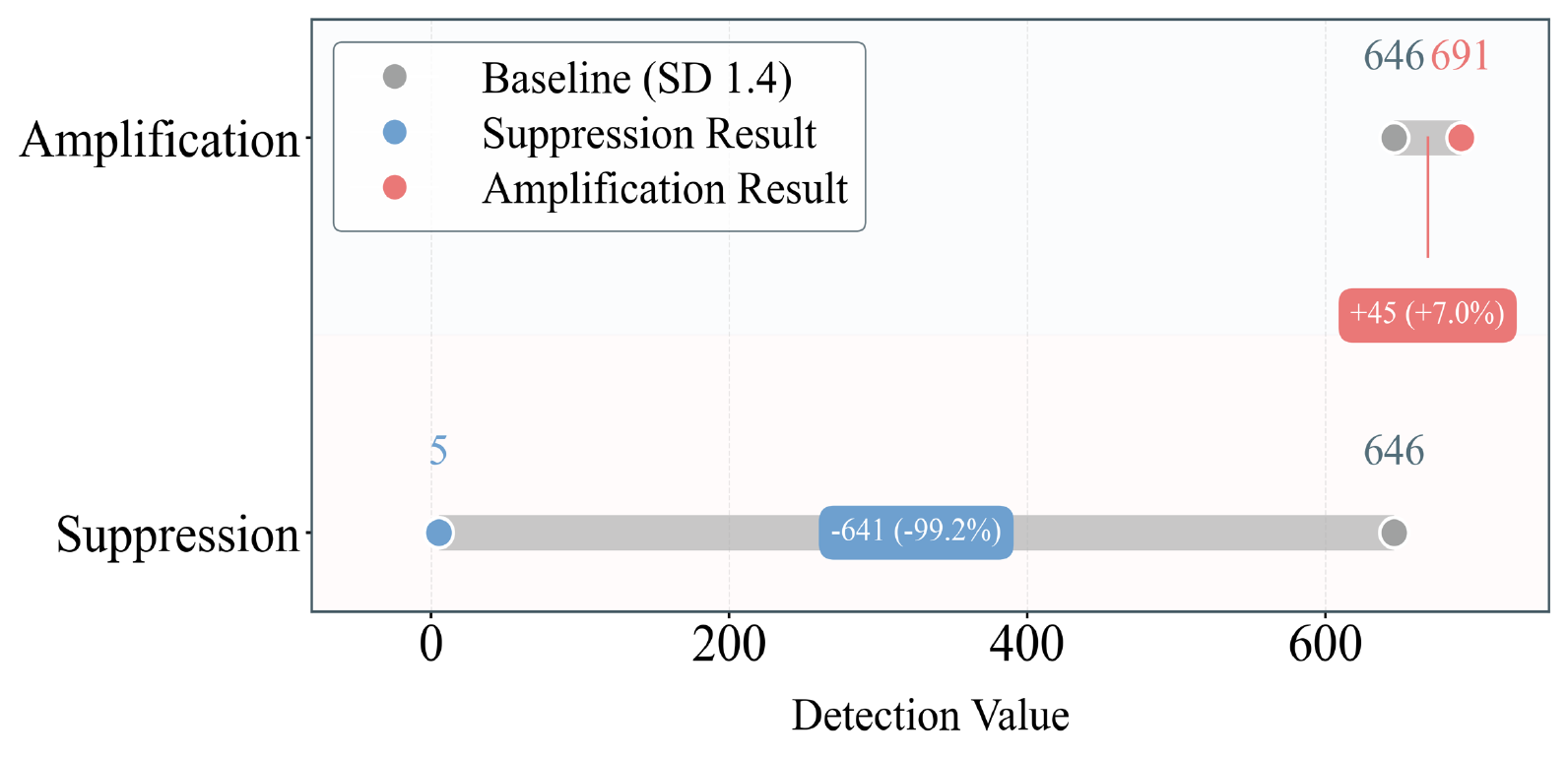}
    \caption{\textbf{Causal validation.} "Amplification" refers to increasing the activation values of neurons.}
    \label{fig:amf_abl}
  \end{minipage}
\end{figure}

% \begin{figure}[t] 
% \centering 
% \includegraphics[width=0.8\linewidth]{figure/neuron.pdf} 
% \caption{\textbf{Neuron-level ablation.} Comparison between random neuron suppression and our targeted selection on the I2P dataset.} 
% \label{fig:neuron_abl} 
% \vspace{-0.2cm}
% \end{figure}

\noindent\textbf{Neuron-Level.}
To verify the functional specificity of identified neurons, we compared our targeted set against a random set of the same size (See Fig.~\ref{fig:neuron_abl}). The results show that only precise suppression yields significant erasure. While the baseline yields 646 detections, random suppression reduces this only marginally to 598, likely due to generic structural disruption. In stark contrast, our targeted neurons reduce the number of detections to 5. This validates that the neurons identified by OrthoEraser are the primary latent carriers of sensitive semantics, making their elimination indispensable for precise erasure without degrading the generative manifold.

% \begin{figure}[t] 
% \centering 
% \includegraphics[width=0.8\linewidth]{figure/amf_fixed.pdf} 
% \caption{\textbf{Causal verification through neuron activation manipulation on the I2P dataset.} "Amplification" refers to increasing the activation values of the neurons identified by OrthoEraser.} 
% \label{fig:amf_abl} 
% \vspace{-0.2cm}
% \end{figure}
% 反直觉反向增强是否会让nudity更严重
\noindent\textbf{Causal Validation.}
To verify the direct causal link between the identified neurons and the sensitive semantics, we conducted an activation amplification experiment in which the activation values of the targeted neurons were intentionally boosted (See Fig.~\ref{fig:amf_abl}). The results show that enhancing these specific neurons significantly increases the probability of the model generating sensitive content, further confirming their role as the core carriers of the concept. Specifically, compared with the 5 detections achieved after OrthoEraser erasure and the 646 detections in the baseline model, amplifying activation of the identified key neurons further increases the detection count to 691. Therefore, this reverse-verification experiment provides strong causal evidence that the neurons identified by our method act as critical switches controlling sensitive information, thereby justifying their selection for precise erasure.

\subsection{More Results and Analysis}

\begin{table*}[t]
\centering
\begin{minipage}{0.48\textwidth}
    \centering
    \caption{\textbf{Analysis of feature entanglement.} We evaluate the impact of different geometric intervention strategies on generative fidelity.}
    \label{tab:entanglement_analysis}
    \resizebox{\textwidth}{!}{
    \begin{tabular}{lccc}
    \toprule
    \textbf{Intervention Strategy} & \textbf{I2P Det. $\downarrow$} &\textbf{CS $\uparrow$} & \textbf{FID $\downarrow$} \\ 
    \midrule
    \color{gray!90} SD1.4 (Baseline) &\color{gray!90}646 & \color{gray!90}31.34 & \color{gray!90}-  \\ 
    \hdashline
    Direct Sensitive Suppression & 17 {\scriptsize ($-629$)}& 30.87 {\scriptsize ($-0.47$)} & 16.64  \\
    Coupled-Aligned Suppression & 12 {\scriptsize ($-634$)} & 26.31 {\scriptsize ($-5.03$)} & 23.95  \\
    \midrule
    \rowcolor[HTML]{ffffb3}\textbf{OrthoEraser (Ours)} & 5 {\scriptsize ($-641$)} & \textbf{31.33} {\scriptsize ($-0.01$)} & \textbf{1.15} \\
    \bottomrule
    \end{tabular}}
\end{minipage}
\hfill
\begin{minipage}{0.48\textwidth}
    \centering
    \caption{\textbf{Ablation study on intervention strategies.} We demonstrate the trade-off between erasure effectiveness and manifold preservation.}
    \label{tab:intervention_ablation}
    \resizebox{\textwidth}{!}{
    \begin{tabular}{lccc}
    \toprule
    \textbf{Intervention Strategy} & \textbf{I2P Det. $\downarrow$} & \textbf{CS $\uparrow$} & \textbf{FID $\downarrow$} \\ 
    \midrule
    \color{gray!90} SD1.4 (Baseline) & \color{gray!90} 646 & \color{gray!90} 31.34 & \color{gray!90} -  \\ 
    \hdashline
    Only Sensitive Suppression & 17 {\scriptsize ($-629$)} & 30.87 {\scriptsize ($-0.47$)} & 16.64  \\ 
    Only Coupled Suppression & 604 {\scriptsize ($-42$)} & 29.95 {\scriptsize ($-1.39$)} & 25.86 \\ 
    \midrule
    \rowcolor[HTML]{ffffb3} \textbf{OrthoEraser (Ours)} & \textbf{5} {\scriptsize ($-641$)} & \textbf{31.33} {\scriptsize ($-0.01$)} & \textbf{1.15} \\ 
    \bottomrule
    \end{tabular}}
\end{minipage}
\end{table*}

\begin{figure}[t] 
\centering 
\includegraphics[width=1\linewidth]{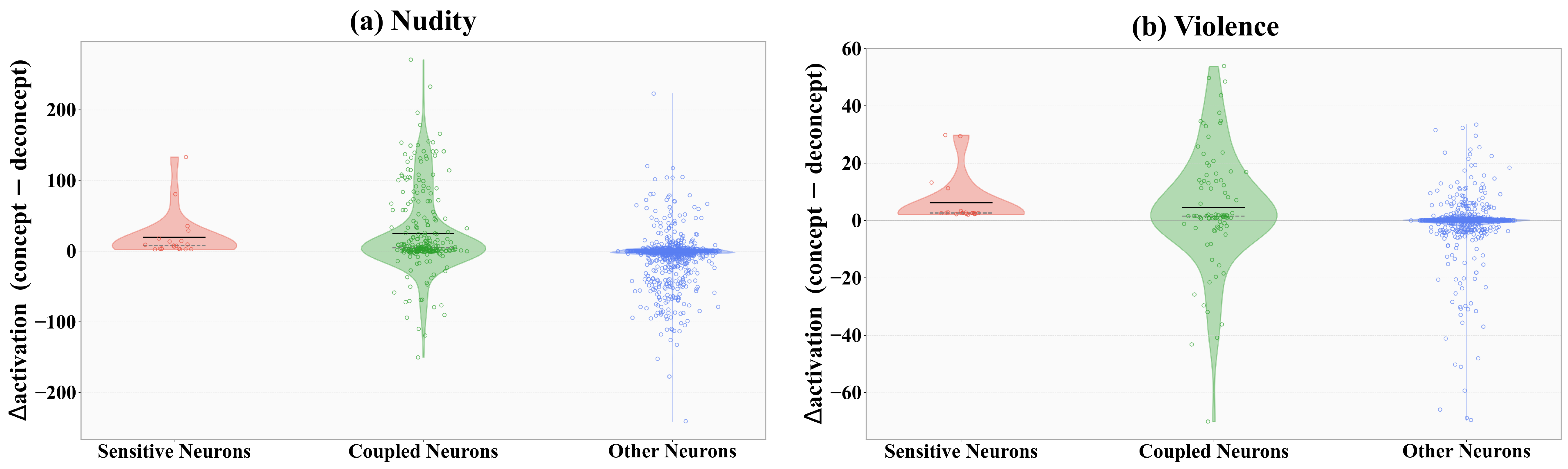} 
\caption{\textbf{Visualizing feature entanglement.} We visualize the distribution of neuron activation shifts ($\Delta activation$) across three categories.} 
\label{fig:jiuchan} 
\end{figure}

\begin{table}[t]
\centering
\caption{\textbf{Quantitative evaluation of generalization (I2P-Violence) and adversarial robustness (Ring-A-Bell, P4D).} Best results are highlighted in \textbf{bold}.}
\label{tab:more}
\renewcommand{\arraystretch}{0.9}
\begin{tabular}{lccc}
\toprule
& \multicolumn{1}{c}{\textbf{Standard Safety (\%)}} & \multicolumn{2}{c}{\textbf{Adversarial Robustness (\%)}} \\
\cmidrule(lr){2-2} \cmidrule(lr){3-4}
\textbf{Method} 
& I2P Violence $\downarrow$
& P4D $\downarrow$
& Ring-A-Bell $\downarrow$
 \\
\midrule
\color{gray!80}SD 1.4~\cite{rombach2022high} & \color{gray!80}40.1 & \color{gray!80}98.7 & \color{gray!80}83.1  \\
\addlinespace[0.5ex] % 增加上方间距
\hdashline
\addlinespace[0.5ex] % 增加上方间距
ESD~\cite{gandikota2023erasing} 
& 16.7 {\tiny (-23.4)} 
& 63.3 {\tiny (-35.4)} 
& 69.7 {\tiny (-13.4)} 
\\

UCE~\cite{gandikota2024unified} 
& 23.3 {\tiny (-16.8)} 
& 80.2 {\tiny (-18.5)} 
& 33.1 {\tiny (-50.0)} 
 \\

% \rowcolor{c4}
% CA~\cite{kumari2023ablating} 
% & 89.8 {\tiny (+7.6)} 
% & -- & -- & -- 
% & 31.21 {\tiny (-0.13)} 
% & 21.55 \\

SLD-Med~\cite{schramowski2023safe} 
& 19.7 {\tiny (-20.4)} 
& 77.5 {\tiny (-21.2)} 
& 66.2 {\tiny (-16.9)} \\

% MACE~\cite{lu2024mace} 
% & 89.1 {\tiny (+6.9)} 
% & -- & -- & -- 
% & 29.32 {\tiny (-2.02)} 
% & 23.45 \\

RECE~\cite{gong2024reliable} 
& 14.2 {\tiny (-25.9)} 
& 64.7 {\tiny (-34.0)} 
& 13.4 {\tiny (-69.7)} 
\\

SPM~\cite{lyu2024one} 
& -- 
& 80.8 {\tiny (-17.9)} 
& 34.2 {\tiny (-48.9)} 
\\

% \rowcolor{c4}
% DuMo~\cite{han2025dumo} 
% & 96.3 {\tiny (+14.1)} 
% & -- & -- & -- 
% & 30.87 {\tiny (-0.47)} 
% & -- \\

SNCE~\cite{he2025single} 
& 17.7 {\tiny (-22.4)} 
& 42.6 {\tiny (-56.1)} 
& 6.3 {\tiny (-76.8)}  \\

\midrule
\rowcolor[HTML]{ffffb3}\textbf{OrthoEraser}
& \textbf{15.6} {\tiny \textbf{(-24.5)}} 
& \textbf{34.6} {\tiny \textbf{(-64.1)}} 
& \textbf{2.7} {\tiny \textbf{(-80.4)}} 
\\
\bottomrule
\end{tabular}
\end{table}

\noindent\textbf{Feature Entanglement.}
To quantify internal entanglement, we perform a neuron activation-shift analysis (Fig.~\ref{fig:jiuchan}). \textcolor[HTML]{EA786C}{Sensitive neurons} exhibit predominantly positive $\Delta$activation, indicating they carry the target harmful concept. \textcolor[HTML]{32A233}{Coupled neurons} show large-magnitude, long-tailed shifts, revealing substantial collateral perturbations on benign semantics, while \textcolor[HTML]{597FF1}{other neurons} remain tightly centered near zero. These observations imply that naive suppression is prone to collateral damage, motivating analytical orthogonalization projection to enforce interventions orthogonal to the protected benign subspace.
We further compare geometric intervention strategies in Table~\ref{tab:entanglement_analysis}. Direct sensitive suppression degrades shared semantics (FID 16.64), and coupled-aligned suppression collapses both fidelity (FID 23.95) and alignment (CS 26.31). Ablations in Table~\ref{tab:intervention_ablation} confirm the failure modes: suppressing only sensitive units leaves residual detections (17), while suppressing only coupled units severely harms the manifold (FID 25.86) yet fails to remove harmful content (604 detections). In contrast, OrthoEraser achieves strong fidelity and alignment (FID 1.15, CS 31.33) by analytically isolating the sensitive direction and projecting it onto the null space of benign features, enabling precise erasure under entanglement.

\begin{figure}[t] 
\centering 
\includegraphics[width=1\linewidth]{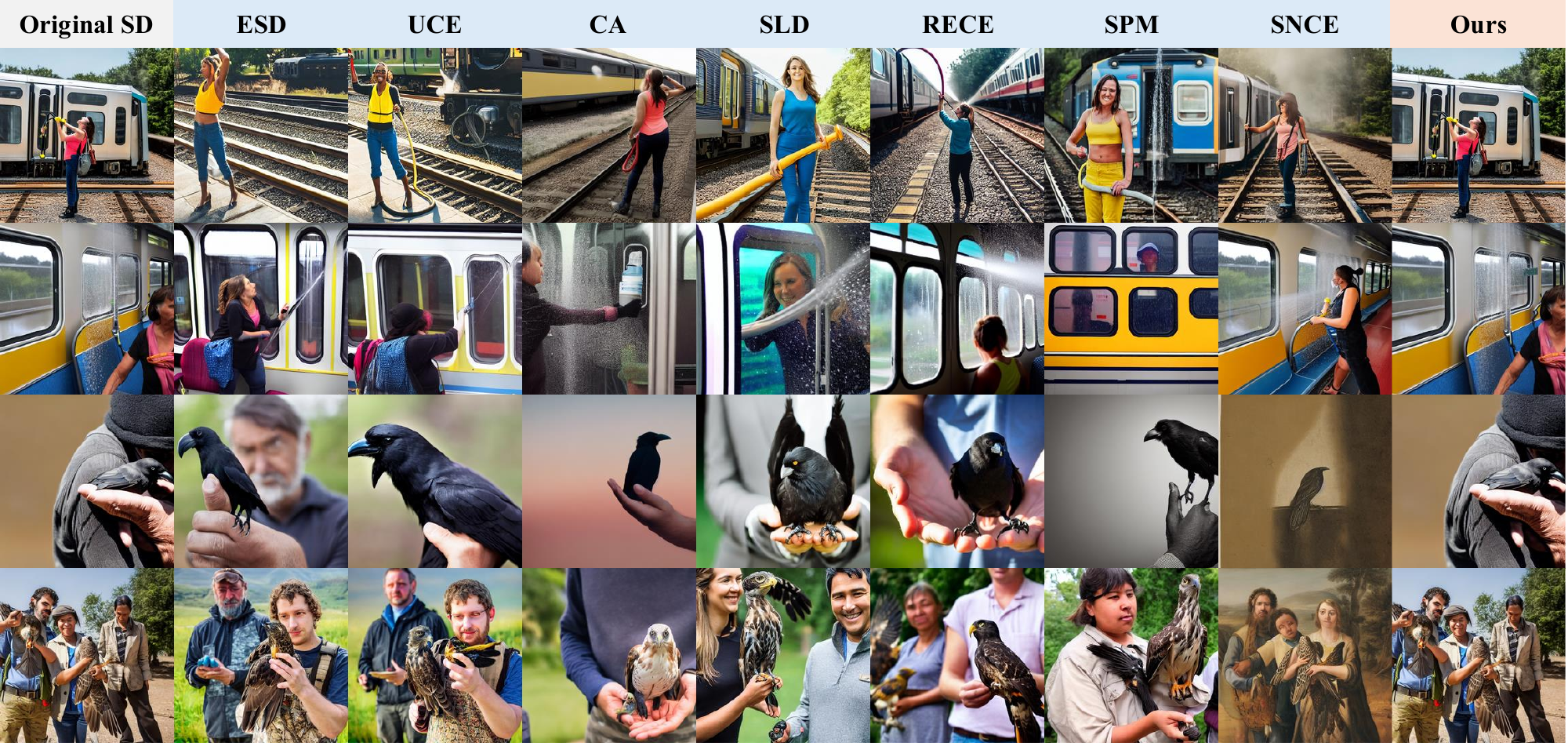} 
\caption{\textbf{Qualitative results on the MS COCO.} We compare the image generation performance of OrthoEraser with other baseline methods using normal prompts.} 
\label{fig:coco} 
\end{figure}

\begin{figure}[t] 
\centering 
\includegraphics[width=1\linewidth]{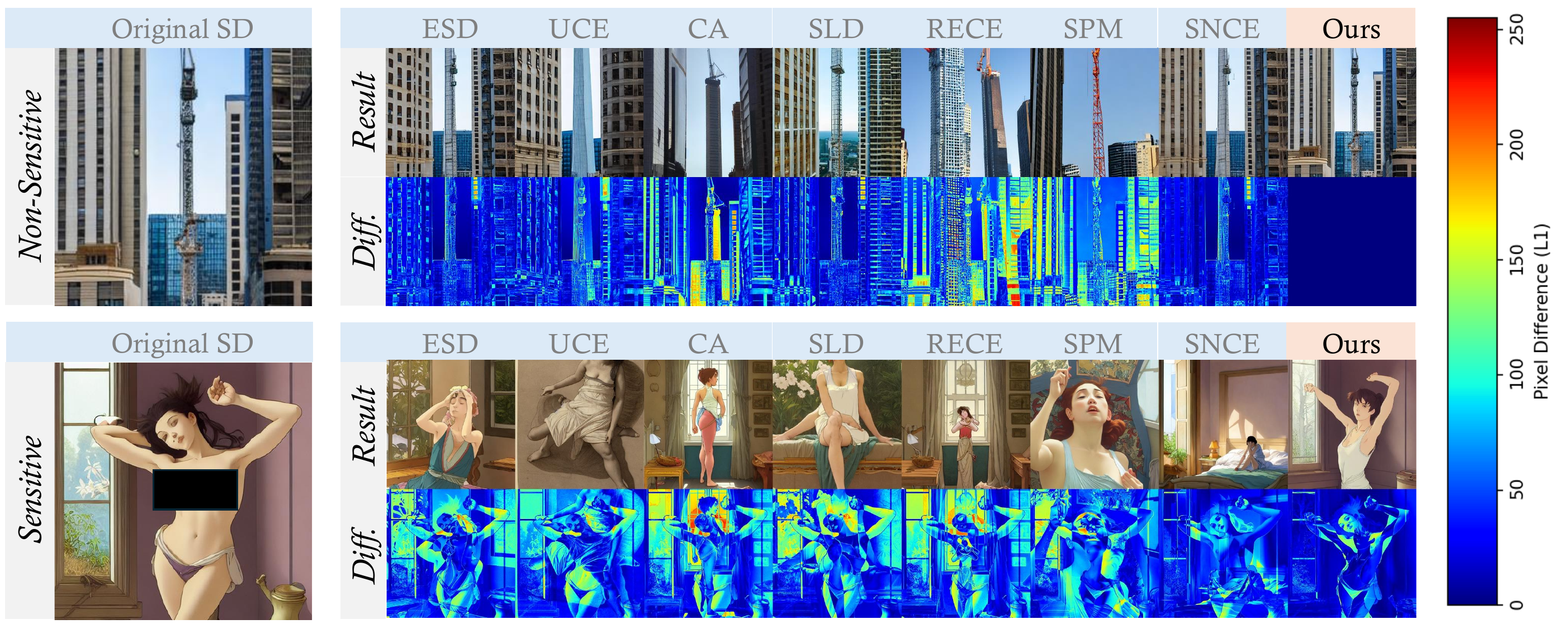} 
\caption{\textbf{Visualizing precision via residuals.} The heatmaps illustrate the intensity of pixel shifts during the concept erasure process for both safe prompts (Top) and sensitive prompts (Bottom).} 
\label{fig:chayi} 
\end{figure}

\noindent\textbf{Benign Fidelity.}
To assess whether erasure harms general generation, we evaluate image quality on MS-COCO (Fig.~\ref{fig:coco}). Due to entanglement, prior methods often introduce collateral artifacts under benign prompts, including semantic drift, structural distortion, and texture degradation. In contrast, OrthoEraser preserves the original model’s fidelity, fine-grained details, and text alignment, consistent with the stable FID and CLIP scores in Table~\ref{tab:nudity_detection}.
We further visualize collateral damage using pixel-wise difference maps (Fig.~\ref{fig:chayi}). For benign prompts (top row), baseline residuals spread broadly and blur normal structures, while OrthoEraser yields near-zero differences. For sensitive prompts (bottom row), baseline edits spill into backgrounds and non-target content, whereas OrthoEraser localizes changes to the targeted sensitive regions and preserves surrounding attributes (e.g., faces and background textures). Together, these results indicate that analytical orthogonalization projection enables precise, high-fidelity safety intervention with minimal collateral damage.

\noindent\textbf{Generalization and Adversarial Robustness.}
To verify the framework's generalization beyond nudity, we performed quantitative evaluations on the I2P-Violence dataset (See Table~\ref{tab:more} Standard Safety). The results indicate that OrthoEraser effectively extends to non-sexual unsafe concepts, consistently achieving superior safety metrics. Specifically, our method reduced the violence detection rate from 40.1\% (SD 1.4) at the baseline to 15.6\%, surpassing SOTA baselines such as ESD (16.7\%) and SNCE (17.7\%). This confirms that analytical orthogonalization projection serves as a versatile framework for eliminating diverse harmful semantics across different modalities.
Then, to evaluate the model's robustness against malicious adversarial attacks and jailbreak prompts, we set up attack success rate (ASR) tests on two challenging adversarial datasets, Ring-A-Bell and P4D (See Table~\ref{tab:more} Adversarial Robustness). The results show that OrthoEraser exhibits superior resistance to adversarial induction, maintaining high safety standards even under aggressive attacks. Specifically, on the highly challenging Ring-A-Bell benchmark, our method reduced ASR from 98.7\% to 2.7\%, and on P4D from 83.1\% to 34.6\%. Therefore, by mathematically severing the activation paths of sensitive concepts via geometric projection, OrthoEraser ensures exceptional robustness against adversarial manipulation.

\begin{figure}[t] 
\centering 
\includegraphics[width=0.9\linewidth]{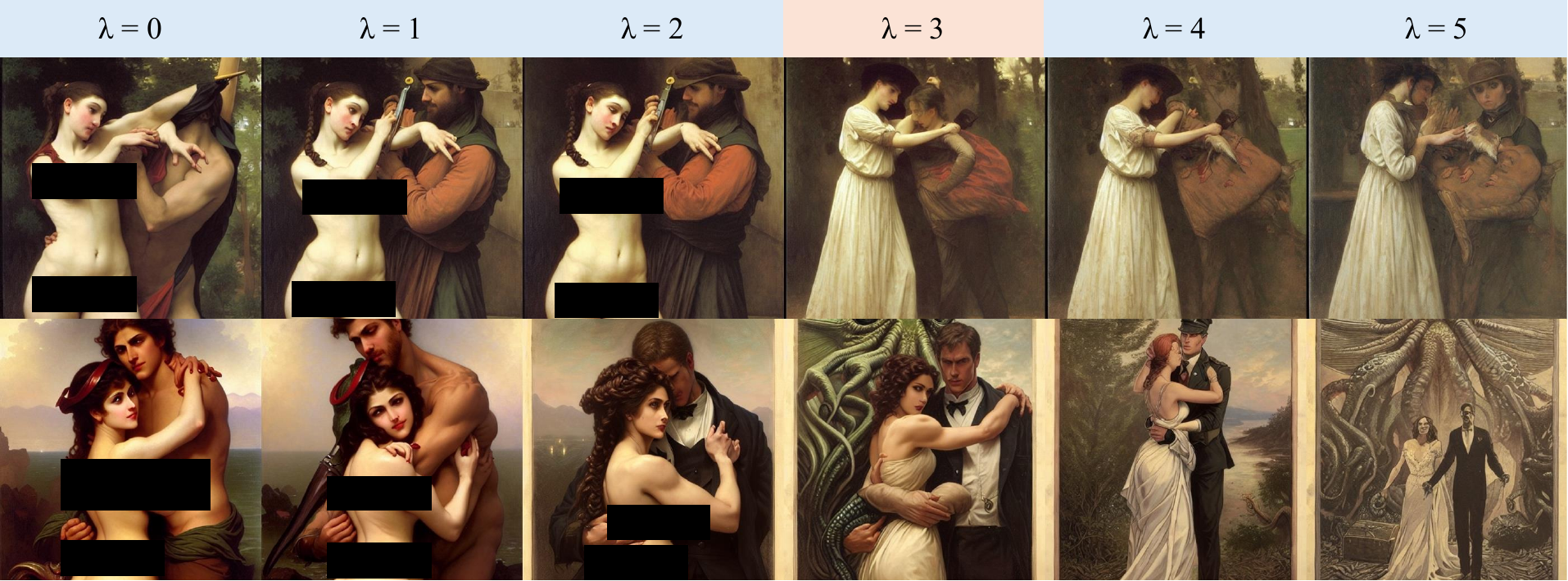} 
\caption{\textbf{Visual ablation of scaling factor $\lambda$.} We illustrate the impact of varying the path-level suppression factor $\lambda$.} 
\label{fig:lambda} 
\end{figure}

\noindent\textbf{Hyperparameter Sensitivity.}
% lambda的变化下定量和定性的分析
To investigate the impact of the scaling factor $\lambda$ on the trade-off between erasure completeness and semantic preservation, we conducted a sensitivity analysis by varying $\lambda$ while monitoring safety performance and image fidelity (See Fig.\ref{fig:lambda}). The results demonstrate a distinct threshold effect, where $\lambda$ governs the critical transition from "under-erasure" to "semantic drift." Specifically, when $\lambda < 3$, the magnitude of the projected intervention vector is insufficient to fully counteract sensitive activations, resulting in residual leakage of harmful content; conversely, once $\lambda$ reaches 3, the erasure effect stabilizes and becomes comprehensive. However, as $\lambda$ increases beyond this optimal point, we observe a noticeable semantic drift in the generated images, indicating that excessive intervention strength begins to perturb the generative manifold. Therefore, we select $\lambda=3$ as the optimal equilibrium, achieving robust erasure while avoiding unnecessary capability loss.

\begin{figure}[t] 
\centering 
\includegraphics[width=0.97\linewidth]{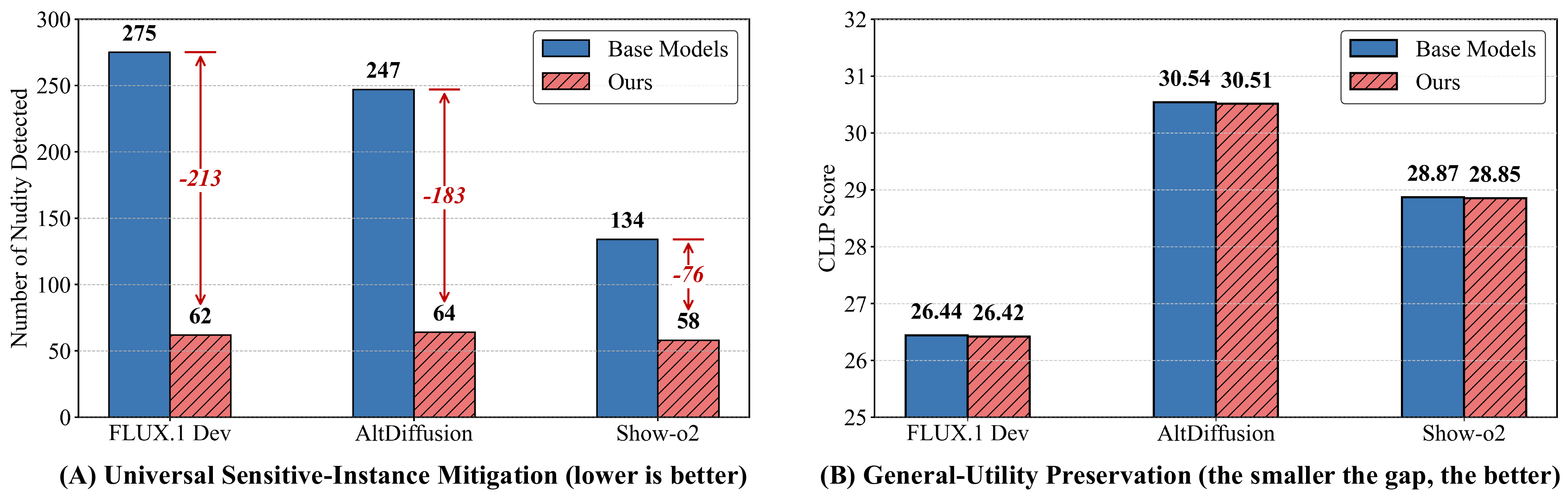} 
\caption{\textbf{Cross-Model Universality.} The left panel shows the number of detected instances for specific body parts, while the right panel evaluates whether the model's general capabilities degrade after neuron correction.} 
\label{fig:fanhua} 
\end{figure}

\noindent\textbf{Cross-Model Universality.}
% 更多生成模型的泛化
To verify universality, we evaluated OrthoEraser across diverse architectures, including the flow-matching-based FLUX.1 Dev, multilingual AltDiffusion, and multimodal Show-o2 (See Fig.\ref{fig:fanhua}). Despite structural differences, all models exhibited significant safety vulnerabilities that our method effectively addressed. Specifically, OrthoEraser reduced nudity detections by 213 (FLUX.1 Dev), 183 (AltDiffusion), and 76 (Show-o2). Crucially, the impact on general utility was negligible, with CLIP Score decreases ranging only from 0.02 to 0.03. These results demonstrate that OrthoEraser is a robust, architecture-agnostic solution for safe foundation model alignment.

\section{Conclusion}
This paper presents OrthoEraser, a concept erasure framework that reduces collateral damage via analytical orthogonalization projection. 
By using sparse autoencoders (SAEs) to localize sensitive directions and constructing a protected subspace from coupled benign features, OrthoEraser enforces orthogonality between the suppressed and the preserved semantic manifolds, enabling precise erasure under feature disentanglement. 
Extensive experiments on several datasets show that OrthoEraser consistently achieves a better safety--fidelity trade-off than prior baselines, delivering strong safety gains while maintaining image quality, text alignment, and an output distribution close to the base model on benign prompts.
It is also observed that OrthoEraser localizes modifications to target regions rather than inducing global semantic drift.

%Bibliography
\bibliographystyle{unsrt}  
\bibliography{references}  

\newpage
\appendix

\section*{Supplementary Material}

The appendices provide supplementary material and theoretical foundations that support the main paper's findings. Appendix ~\ref{th} delivers a formal theoretical justification for the OrthoEraser framework, including a closed-form mathematical derivation of the analytical null-space projection via the method of Lagrange multipliers and a computational complexity analysis. Appendix ~\ref{discussion} provides an extended discussion addressing critical aspects of the framework, such as the scalability constraints in dense multi-concept erasure, the potential for adaptive thresholding, and the challenges of intervening across distributed semantic representations. Appendix ~\ref{limitation} details the limitations of the current approach and outlines future work, specifically analyzing the framework's inherent dependence on the dictionary capacity, resolution, and feature disentanglement quality of the underlying sparse autoencoders (SAEs).

\section{Theoretical Analysis}
\label{th}

In this section, we provide a formal mathematical derivation for the gradient orthogonalization strategy proposed in OrthoEraser. We demonstrate that our method is the closed-form solution to a constrained optimization problem that minimizes linear interference with the semantic subspace of coupled neurons while maximizing the erasure of sensitive concepts.

\subsection{Problem Formulation}

Let $h \in \mathbb{R}^d$ be the dense latent activation vector at the identified sensitive layer $l^*$. Our goal is to find an optimal intervention direction $d^*$ to subtract from $h$, yielding a safe latent representation $\tilde{h} = h - \lambda d^*$, where $\lambda$ is the scalar suppression strength.

We can formally define two opposing objectives for this intervention direction. First is the \textbf{Erasure Objective}: the direction $d^*$ should align as closely as possible with the raw sensitive direction $d_{raw}$, where $d_{raw} = \sum_{i \in \mathcal{N}_{sens}} z_i w_i^{dec}$. Second is the \textbf{Preservation Constraint}: the intervention must not perturb the structurally entangled benign concepts. Specifically, for the set of coupled neurons $\mathcal{C}$ identified via zero-ablation (Sec. 3.2), the projection of the new latent state $\tilde{h}$ onto their corresponding decoder weights $W_{\mathcal{C}} \in \mathbb{R}^{d \times |\mathcal{C}|}$ must identically match the projection of the original latent state $h$.

Mathematically, the preservation constraint requires that:
\begin{equation}
    W_{\mathcal{C}}^\top \tilde{h} = W_{\mathcal{C}}^\top h
\end{equation}

Substituting $\tilde{h} = h - \lambda d^*$, we obtain:
\begin{equation}
    W_{\mathcal{C}}^\top (h - \lambda d^*) = W_{\mathcal{C}}^\top h \implies \lambda W_{\mathcal{C}}^\top d^* = 0
\end{equation}

Assuming a non-zero intervention strength ($\lambda > 0$), this simplifies to the geometric requirement that the intervention direction must be orthogonal to the coupled benign features: $W_{\mathcal{C}}^\top d^* = 0$. We formulate the search for the optimal concept erasure direction as the following constrained least-squares optimization problem:
\begin{equation}
    d^* = \arg\min_{d \in \mathbb{R}^d, \; W_{\mathcal{C}}^\top d = 0} \frac{1}{2} \| d - d_{raw} \|_2^2.
\end{equation}

\subsection{Closed-Form Derivation via Method of Lagrange Multipliers}

To solve this optimization problem, we introduce a vector of Lagrange multipliers $\nu \in \mathbb{R}^{|\mathcal{C}|}$ and define the Lagrangian functional $\mathcal{L}$:
\begin{equation}
    \mathcal{L}(d, \nu) = \frac{1}{2} \| d - d_{raw} \|_2^2 + \nu^\top W_{\mathcal{C}}^\top d.
\end{equation}

To find the minimum, we set the gradient of $\mathcal{L}$ with respect to the direction $d$ to zero:
\begin{equation}
    \nabla_{d} \mathcal{L} = (d - d_{raw}) + W_{\mathcal{C}} \nu = 0,
\end{equation}
\begin{equation}
    d = d_{raw} - W_{\mathcal{C}} \nu.
\end{equation}

To find $\nu$, we substitute this expression back into our orthogonality constraint ($W_{\mathcal{C}}^\top d = 0$):
\begin{equation}
    W_{\mathcal{C}}^\top (d_{raw} - W_{\mathcal{C}} \nu) = 0,
\end{equation}
\begin{equation}
    W_{\mathcal{C}}^\top W_{\mathcal{C}} \nu = W_{\mathcal{C}}^\top d_{raw}.
\end{equation}

Assuming the coupled decoder weights in $W_{\mathcal{C}}$ are linearly independent (a standard property of sparse bases learned by SAEs), the Gram matrix $(W_{\mathcal{C}}^\top W_{\mathcal{C}})$ is invertible. We then solve for $\nu$:
\begin{equation}
    \nu = (W_{\mathcal{C}}^\top W_{\mathcal{C}})^{-1} W_{\mathcal{C}}^\top d_{raw}.
\end{equation}

Substituting the optimal $\nu$ back into the equation for the optimal direction $d^*$:
\begin{align}
    d^* &= d_{raw} - W_{\mathcal{C}} \left( (W_{\mathcal{C}}^\top W_{\mathcal{C}})^{-1} W_{\mathcal{C}}^\top d_{raw} \right) \nonumber \\
    &= \left[ I - W_{\mathcal{C}} (W_{\mathcal{C}}^\top W_{\mathcal{C}})^{-1} W_{\mathcal{C}}^\top \right] d_{raw}.
\end{align}

\subsection{Equivalence to OrthoEraser's QR Projection}

We now demonstrate that the bracketed term is mathematically equivalent to the projection mechanism utilized in OrthoEraser. Let us define the matrix $P_{full}$ as:
\begin{equation}
    P_{full} = W_{\mathcal{C}} (W_{\mathcal{C}}^\top W_{\mathcal{C}})^{-1} W_{\mathcal{C}}^\top.
\end{equation}
By definition, $P_{full}$ is the standard orthogonal projection matrix onto the column space of $W_{\mathcal{C}}$.

In Sec 3.3, OrthoEraser computes an orthonormal basis $Q$ for the column space of $W_{\mathcal{C}}$ via QR decomposition: $W_{\mathcal{C}} = QR$, where $Q^\top Q = I$. Substituting $QR$ for $W_{\mathcal{C}}$ in the definition of $P_{full}$:
\begin{align}
    P_{full} &= (QR) ((QR)^\top QR)^{-1} (QR)^\top \nonumber \\
    &= QR (R^\top Q^\top QR)^{-1} R^\top Q^\top \nonumber \\
    &= QR (R^\top R)^{-1} R^\top Q^\top \nonumber \\
    &= QR R^{-1} (R^\top)^{-1} R^\top Q^\top \nonumber \\
    &= Q (I) (I) Q^\top = QQ^\top.
\end{align}

This strictly matches our definition of $P = QQ^\top$. Therefore, the optimal intervention vector $\Delta h$ simplifies to:
\begin{equation}
    \Delta h = -\lambda (I - P) d_{raw}.
\end{equation}

Defining the orthogonalized pure sensitive direction as $d^* = (I - P) d_{raw}$, we arrive at the final update rule:
\begin{equation}
    \tilde{h} = h + \Delta h = h - \lambda d^*.
\end{equation}

This derivation proves that OrthoEraser's update rule is the exact closed-form solution to minimizing sensitive features while theoretically eliminating linear interference with the coupled benign semantics.

\subsection{Theoretical Assumptions and Boundaries}

While the closed-form solution provides a rigorous mathematical foundation for OrthoEraser, it is crucial to articulate the foundational assumptions that bound this theoretical optimality. Our formulation operates under three primary approximations. 

\noindent\textbf{Local Linear Approximation of Non-linear Manifolds.} The constraint $W_{\mathcal{C}}^\top \Delta h = 0$ eliminates interference strictly within the linear subspace spanned by the coupled neurons $\mathcal{C}$ at the specific intervention layer $l^*$. Because T2I models consist of deep, highly non-linear transformations across subsequent layers, perfectly decoupling concepts at the final image-pixel level is analytically intractable. Therefore, our null-space projection should be interpreted as a first-order linear approximation of manifold preservation. By geometrically severing the activation path at the most sensitive layer, we effectively minimize the downstream propagation of the intervention signal.

\noindent\textbf{Linear Independence of Sparse Basis.} The derivation relies on the invertibility of the Gram matrix $(W_{\mathcal{C}}^\top W_{\mathcal{C}})$. This assumes that the decoder weight vectors of the coupled benign neurons $W_{\mathcal{C}}$ are linearly independent. In practice, SAEs naturally encourage near-orthogonal or linearly independent bases for highly active features. If multi-collinearity were to occur among the coupled neurons, the Moore-Penrose pseudo-inverse $(W_{\mathcal{C}}^\top W_{\mathcal{C}})^+$ could be seamlessly substituted to compute the projection matrix $P$, ensuring the optimization problem remains well-posed.

\noindent\textbf{SAE Subspace Completeness.} We operate under the assumption that the SAE's high-dimensional latent space captures a complete representation of both the sensitive and coupled benign semantics. Any residual sensitive information encoded in the SAE's reconstruction error resides outside this parameterized basis and is therefore bypassed by the projection matrix. Our layer-wise sensitive selection implicitly mitigates this by locating the layer where the concepts are maximally disentangled.

\subsection{Computational Complexity Analysis}

A key advantage of OrthoEraser is its minimal impact on inference speed. The additional computational overhead is concentrated strictly at the target intervention layer $l^*$. Given a dense representation of dimension $d$ and a coupled neuron set of size $|\mathcal{C}|$, the computational bottlenecks consist of three operations. First, encoding the dense features into the sparse space incurs an SAE Inference complexity of $\mathcal{O}(d \cdot D_{SAE})$, where $D_{SAE}$ is the SAE hidden dimension. Second, computing $W_{\mathcal{C}} = QR$ on a matrix of computing $W_{\mathcal{C}} = QR$ on a matrix of size $d \times |\mathcal{C}|$ incurs a one-time offline cost of $\mathcal{O}(d \cdot |\mathcal{C}|^2)$ during the detection phase. During inference, the primary overhead is the orthogonal projection, which involves a matrix-vector multiplication $QQ^\top d_{raw}$ with a complexity of $\mathcal{O}(d \cdot |\mathcal{C}|)$. Since the number of coupled neurons $|\mathcal{C}|$ is typically much smaller than the latent dimension $d$ (i.e., $|\mathcal{C}| \ll d$), this overhead is negligible compared to the billions of operations in a standard Diffusion U-Net or Transformer forward pass. Consequently, OrthoEraser achieves precise safety alignment with near-zero latency impact, making it suitable for real-time generative applications.

\section{More Discussions}
\label{discussion}
\noindent$\triangleright$ \textbf{\textit{Q1. Why does the method rely heavily on SAEs, and how does SAE quality affect the erasure of highly abstract concepts?}}

A fundamental premise of our framework is the use of SAEs for high-resolution feature disentanglement. Consequently, the precision of our geometric projection is inherently limited by the SAE's dictionary size and reconstruction fidelity. If an exceptionally abstract concept is not explicitly captured by the sparse basis, our targeted erasure may experience slight semantic drift. We respectfully view this as an infrastructure dependency rather than an algorithmic flaw. As the community continues to scale SAE architectures, our analytical framework will seamlessly inherit these representational improvements without requiring modifications to the core optimization algorithm.

\noindent$\triangleright$ \textbf{\textit{Q2. Why does OrthoEraser intervene at only a single layer, and is this sufficient for unlearning complex concepts distributed across multiple layers?}}

OrthoEraser intervenes at the single most critical layer exhibiting maximum attention divergence. While this surgical projection is computationally efficient and empirically sufficient to block explicitly harmful content, we humbly recognize that deep generative models process information hierarchically. Certain deeply ingrained biases or nuanced artistic styles may be distributed across multiple attention layers. Investigating a cascaded null-space projection that coordinates orthogonal constraints across a sequential multi-layer pathway is an exciting and necessary open challenge for achieving truly comprehensive concept unlearning.

\noindent$\triangleright$ \textbf{\textit{Q3. Why does the framework use a fixed Top-k selection for coupled neurons, and how might this limit the preservation of benign semantics across diverse prompts?}}

We currently define the protected coupled set using a fixed top-k threshold based on zero-ablation activation shifts. While our evaluations demonstrate that this effectively balances erasure precision and general utility, we acknowledge that semantic complexity varies significantly across different prompts. For instance, a highly detailed compositional prompt naturally entangles more benign features than a simple portrait. Future iterations could greatly benefit from exploring dynamic, adaptive thresholding mechanisms based on real-time activation distributions, offering a more mathematically optimal preservation of benign semantics on a strictly case-by-case basis.

\noindent$\triangleright$ \textbf{\textit{Q4. Why might the analytical null-space projection struggle with mass concept erasure, and what geometric bottlenecks arise when erasing hundreds of concepts?}}

OrthoEraser enforces interventions to be strictly orthogonal to the identified benign subspace. However, in demanding scenarios like simultaneous mass copyright erasure, the size of the protected coupled set inevitably expands. From a linear algebra perspective, this continuous expansion increases the rank of our projection matrix, progressively narrowing the available null space. In extreme cases, the optimization space becomes over-constrained, reducing the intervention vector to near-zero. Addressing this geometric bottleneck, perhaps via block-diagonal orthogonalization or soft penalty constraints, is a critical next step for scaling analytical erasure methods to enterprise levels.

\noindent$\triangleright$ \textbf{\textit{Q5. Why is the current evaluation limited strictly to spatial text-to-image (T2I) generation, and what are the challenges in adapting this to temporal or 3D modalities?}}

We deliberately limited our scope to foundational T2I models, where spatial feature entanglement is the primary safety concern. While our analytical null-space projection elegantly addresses this, generative AI is rapidly evolving towards complex temporal (video) and 3D modalities. Applying OrthoEraser to video diffusion models requires not only spatial disentanglement but also tracking and orthogonalizing concept pathways across temporal attention layers to prevent flickering. We consider adapting our geometric constraints to maintain temporal coherence as a highly promising frontier for future safety alignment research.

\section{Limitation}
\label{limitation}
A fundamental premise of OrthoEraser is the utilization of sparse autoencoders (SAEs) to decompose dense, polysemantic activations into a high-resolution, interpretable basis. Consequently, we respectfully note that the precision of our geometric projection is inherently bottlenecked by the resolution, dictionary capacity, and reconstruction fidelity of the underlying SAE. In our implementation, we utilized a top-$k$ SAE with a specific expansion factor; however, if an exceptionally abstract or rare sensitive concept is not explicitly captured by the learned sparse basis, or if it remains stubbornly entangled within a single neuron due to capacity limits, our targeted erasure may experience slight semantic drift. We view this not as an algorithmic flaw of the orthogonal projection itself, but rather as an upstream infrastructure dependency. As the generative AI community continues to scale and refine SAE architectures to achieve true feature monosemanticity, our analytical framework will seamlessly inherit these representational improvements, naturally enhancing its erasure precision without requiring modifications to our core optimization algorithm.

\end{document}